\documentclass[10pt,twocolumn,letterpaper]{article}

\usepackage[pagenumbers]{iccv} 

\definecolor{iccvblue}{rgb}{0.21,0.49,0.74}
\usepackage[pagebackref,breaklinks,colorlinks,allcolors=iccvblue]{hyperref}
\usepackage{multirow}
\title{Follow-Your-Color: Multi-Instance Sketch Colorization}

\author{
Yinhan Zhang\textsuperscript{1\dag} \quad 
Yue Ma\textsuperscript{2\dag} \quad 
Bingyuan Wang\textsuperscript{1} \quad 
Qifeng Chen\textsuperscript{2} \quad 
Zeyu Wang\textsuperscript{1,2*}
\\
\textsuperscript{1}The Hong Kong University of Science and Technology (Guangzhou)\\
\textsuperscript{2}The Hong Kong University of Science and Technology
}

\begin{document}

\twocolumn[{
\renewcommand\twocolumn[1][]{#1}%
\maketitle

\begin{center}
  \centering
  \includegraphics[width=0.85\linewidth]{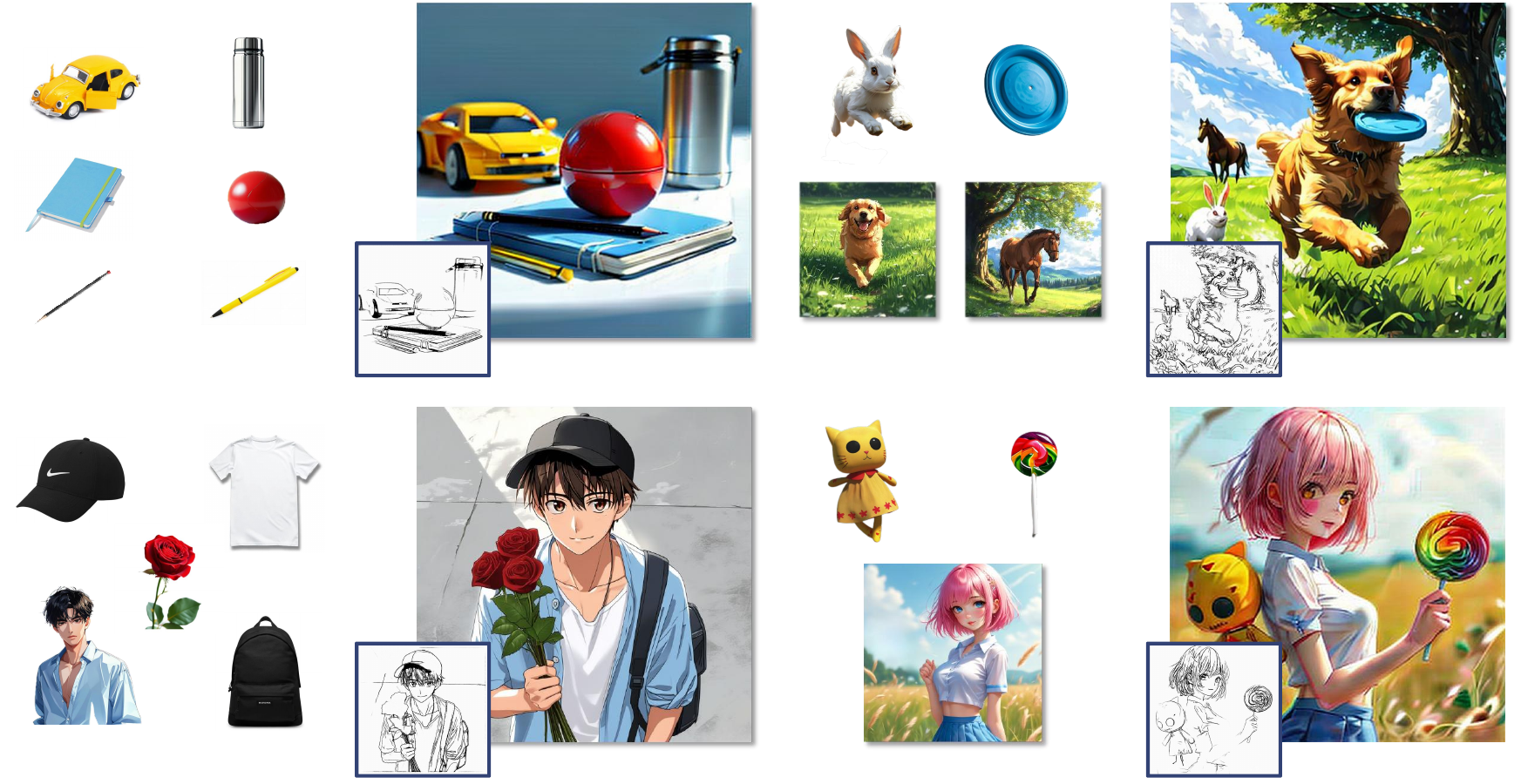}
  \captionof{figure}{\textbf{Results of Follow-Your-Color.} Given a set of colored references, Follow-Your-Color can colorize a line art image while maintaining color consistency across multiple instances. Compared to traditional methods, our approach significantly improves coloring efficiency.
  }
\end{center}
}]

\newcommand\blfootnote[1]{%
  \begingroup
  \renewcommand\thefootnote{}\footnote{#1}%
  \addtocounter{footnote}{-1}%
  \endgroup
}

\blfootnote{$\dagger$ Equal contribution.}
\blfootnote{$*$ Corresponding author.}

\begin{abstract}
We present \textit{Follow-Your-Color}, a diffusion-based framework for multi-instance sketch colorization. 
The production of multi-instance 2D line art colorization adheres to an industry-standard workflow, which consists of three crucial stages: the design of line art characters, the coloring of individual objects, and the refinement process.  The artists are required to repeat the process of coloring each instance one by one, which is inaccurate and inefficient. Meanwhile, current generative methods fail to solve this task due to the challenge of multi-instance pair data collection.
To tackle these challenges, we incorporate three technical designs to ensure precise character detail transcription and achieve multi-instance sketch colorization in a single forward pass.
Specifically, we first propose the self-play training strategy to 
address the lack of training data. Then we introduce an instance guider to feed the color of the instance. To achieve accurate color matching, we present fine-grained color matching with edge loss to enhance visual quality. Equipped with the proposed modules, 
\textit{Follow-Your-Color} enables automatically transforming sketches into vividly-colored images with accurate consistency and multi-instance control.
Experiments on our collected datasets show that our model outperforms existing methods regarding chromatic precision. Specifically, our model critically automates the colorization process with zero manual adjustments, so novice users can produce stylistically consistent artwork by providing reference instances and the original line art. 
Our code and additional details are available at \url{https://yinhan-zhang.github.io/color}.

\end{abstract}
\begin{figure*}[htbp]
  \centering
  \includegraphics[width=\linewidth]{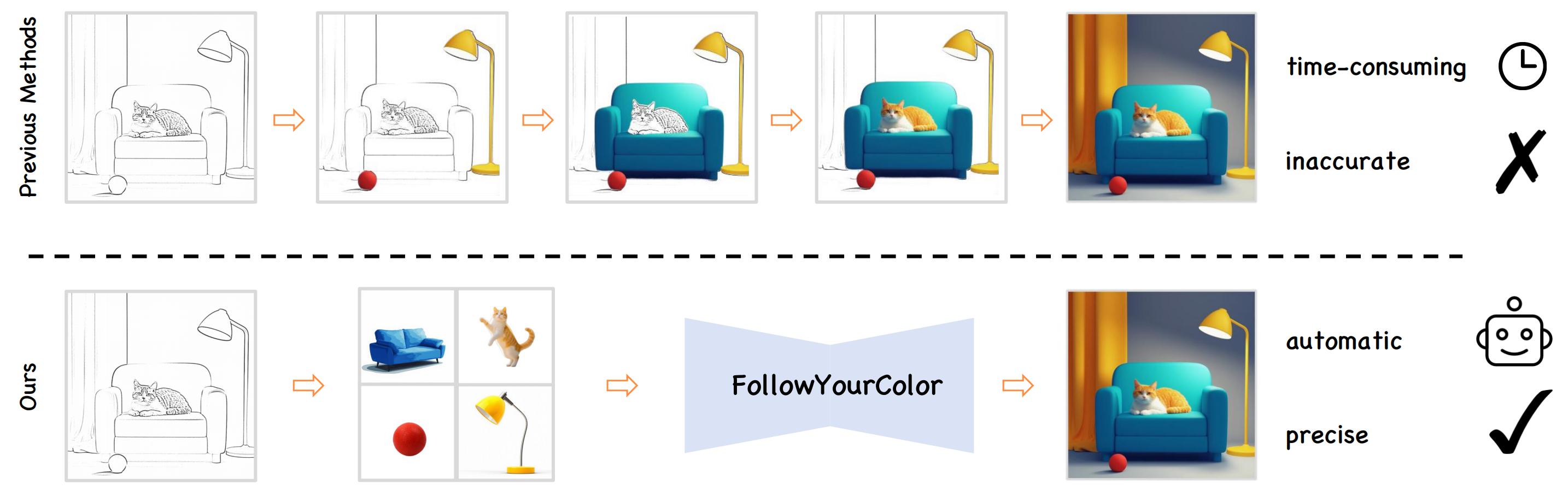}
  \caption{\textbf{Illustration of the workflow of multi-instance sketch colorization production.}
  Previous methods can only achieve multi-instance sketch colorization step by step, which is time-consuming and inaccurate. In contrast, our method can color a sketch while maintaining consistency, making multi-instance sketch colorization easier. }
\end{figure*}

\section{Introduction}

The fast development of the digital cartoon industry has demonstrated widespread potential applications for generative artificial intelligence. Colorizing cartoon sketches, a crucial aspect within the broader realm of the cartoon industry, serves as an indispensable task. It not only heightens the visual allure but also enriches the narrative experience by effectively and vividly communicating emotions and actions. Using the automatic pipeline can streamline production workflows, accelerate content creation, reduce labor demands, and meet the growing demand for cartoons.

In the traditional process of cartoon sketch colorization, artists first analyze the line art to grasp the character and story context. They then choose suitable color palettes and manually apply colors layer by layer with attention to detail. 
However, traditional methods have two drawbacks. The first one is that they are time-consuming, as manual coloring requires great effort to match the original design. For instance, in an animation series, multiple identical objects in the same image need to be colored one by one, which is very inefficient. The other problem is inaccuracy since manual work is error-prone and different artists may have varied color interpretations, causing inconsistent results in large-scale projects. Our method slots into this workflow, supporting auto-colorization while preserving design fidelity and instance-level color consistency. 

Cartoon sketch colorization has garnered considerable research interest due to its potential in digital art and animation production.
Currently, user-guided methods~\cite{Diffusart, Twostagesketch, Colorformer, Scribbler, scrible2, scrible3, Regionmap, Tag2pix, Instructpix2pix} rely on explicit human inputs such as color points, scribbles, or textual descriptions to direct the coloring process. Reference-based approaches~\cite{example1, example2, example3, example4, example5} employ fused attention mechanisms to transfer colors from exemplar images while preserving structural coherence.
While recent advances~\cite{Gan-supervised, Graphmatching} have improved automated colorization, three critical limitations persist:
(1) \textbf{Domain adaptation gap:} existing colorization pipelines rely heavily on reference image fidelity, as pronounced structural mismatches between line art and exemplars frequently induce erroneous chromatic mappings. This dependency imposes an impractical requirement for near-isomorphic geometry between inputs, which rarely holds in unconstrained animation workflows where stylized sketches often diverge significantly from real-world references.
(2) \textbf{Instance-level control granularity:} existing methods demonstrate a lack of fine-grained control over instance-specific attributes during the color transfer process. Discrepancies in character pose, proportion, or viewpoint between reference instances and target sketches often lead to distorted textures and the loss of important details. As a result, crucial features from the reference image may be significantly diminished during the colorization process.
(3) \textbf{Color consistency:} achieving instance-aware color consistency is essential. The coloring of regions with non-closed lines or multiple objects often results in color bleeding, which compromises the accuracy and stability of instance control. This instability undermines the overall harmony of the image, which detracts from the viewer’s experience.

To overcome these challenges, we propose \textit{Follow-Your-Color}, a novel framework that streamlines the colorization process. Our approach builds upon pre-trained diffusion model priors, capitalizing on their learned capacity to enforce visual consistency across generated outputs. The architectural cornerstones of our approach are articulated as follows.
First, to tackle the misalignment between the reference instances and target line art, we incorporate an explicit reference mechanism that seamlessly infuses reference-derived color semantics and artistic styles into line art. In addition, we employ implicit latent control to ensure precise reference to each instance, significantly enhancing color accuracy and consistency.
Second, to further improve the visual quality and color consistency of the output, we introduce edge loss and color matching. These methods compel the model to genuinely extract color information from the reference character design, thereby improving the accuracy of semantic correspondence and reducing the reliance on any color information that may inadvertently leak from the reference image.
Third, our model adopts a two-stage, self-play training strategy, which addresses the challenges of limited multi-instance training data and subsequently incorporates additional reference images to refine the colorization capability. By facilitating colorization across multiple instances, our model achieves impressive color consistency with minimal human intervention.
Our approach achieves state-of-the-art results across both quantitative metrics and qualitative evaluations, outperforming prior art in animated content creation. We aim for a critical step toward fully automated, high-efficiency animation pipelines with guaranteed stylistic coherence. This method can also be extended beyond anime to the broader digital art and media fields. Our contributions can be summarized as follows:
\begin{itemize}
   \item  We propose \textbf{\textit{Follow-Your-Color}}, the first multi-instance coloring method to support multiple instances integration for sketch colorization in a single forward pass.
   
   \item Technically, to solve the lack of multi-instance data, we design a two-stage, self-play training strategy. We also propose an instance guider and pixel-level color matching with edge loss to enhance the color correspondence.

\end{itemize}
\section{Related Work}

\subsection{Line Art Colorization}
Line art colorization techniques strive to decode the link between semantics and color by leveraging large-scale datasets~\cite{automaticanimation, ma2024followyourpose, ma2024followyourclick, ma2022visual, ma2023magicstick, ma2024followyouremoji, DiffusingColors, Animationtransformer, Clic, yuan2024identity, yuan2024magictime}. Researchers utilize a series of semantic modules such as classification~\cite{classification}, semantic segmentation~\cite{Pixcolor, pixelated}, and instance-aware information~\cite{instance-aware, Dreambooth} to enrich color vibrancy. These methods generally perform well when the object's color has a strong semantic-based determinacy. However, when confronted with objects that exhibit a broad range of colors, they often produce lackluster results. In contrast, our proposed framework, equipped with an innovative imagination module, is engineered to transcend this constraint. 

On the other hand, generative priors enshrined within pretrained Generative Adversarial Networks (GANs)~\cite{Gan, Gan2, Cgan} and Diffusion models have been cornerstones in pursuing photorealistic colorization. GANs are adept at both conditional and unconditional image synthesis. Specific conditioning factors, such as layout and semantic maps, are harnessed to fine-tune the image synthesis process~\cite{Layout_style, Layout_style2}. For example, the StyleGAN-family models have demonstrated remarkable prowess in generating high-resolution images without explicit conditioning~\cite{StyleGan, Stylegan2, Cycleconsistency}. Some models struggle to maintain the local spatial integrity of grayscale inputs but can use diverse color priors from pre-trained models ~\cite{Dreambooth, Unicolor, Dualspace}. In our sketch colorization task, we aim to adapt these generative prior-based techniques. Our method can preserve sketch structures and semantics while using their color-generation abilities.

\subsection{Visual Correspondence}
In computer vision, visual correspondence aims to identify and match relevant features or points among different images. It is widely applied in tasks such as stereo vision and motion tracking. In the past, traditional methods~\cite{surf, scale-invariant} relied on hand-designed features to establish corresponding relationships. Nowadays, deep learning methods~\cite{cats, patchmatch, transformatcher, chen2024follow} obtain matching capabilities through supervised learning with labeled datasets. However, supervised learning faces significant scalability issues. Precise pixel-level annotations are not only time-consuming and labor-intensive but also costly. To address this, scholars have started to explore weakly-supervised or self-supervised visual correspondence models. For example, LightGlue~\cite{lightglue} can match sparse local features across image pairs through an adaptive mechanism. CoTracker~\cite{cotracker3} adopts a semi-supervised training method by generating pseudo-labels using off-the-shelf models. DIFT~\cite{Dift} extracts features by diffusion models and can achieve pixel-level semantic point matching. Building upon the diffusion models' prior correspondence knowledge, our framework enables reference-based colorization via semantic-aligned color mapping between line art and reference instances, without any structural modifications.

\subsection{Reference-Based Image Colorization}
A great deal of research has been committed to colorizing photographs using reference-based priors~\cite{Prompt2prompt, Masactrl, Paintbyexample, Textinversion, Realfill}. Initially, efforts focused on the transfer of chromatic information to the corresponding regions through luminance and texture alignment, with various low-level feature-based correspondence techniques developed for more precise local color transfer~\cite{Videocolor, Reference1, Diffusart, wang2024cove, Graphmatching}. However, these methods are vulnerable to complex appearance variations of the same object, as low-level features cannot capture semantic nuances~\cite{Pixcolor, Colorfulimagecolor}. Line art colorization is notably different from natural image colorization~\cite{Structure, Imagequalityassessment, Twostagesketch, Twosteptraining}. 

The Diffusion models have emerged as a powerful alternative. The Denoising Diffusion Probabilistic Model~\cite{Ddpm} and the Denoising Diffusion Implicit Model~\cite{Ddim} paved the way for Latent Diffusion Models (LDMs) like Stable Diffusion~\cite{Ldm}, revolutionizing text-to-image generation. Building on LDMs, ControlNet ~\cite{controlnet} uses task-specific conditions and multi-modal inputs~\cite{Auto-painter, Dreaminpainter, Gan-supervised} to control pre-trained diffusion models~\cite{Flexicon, zhu2024instantswap, feng2024dit4edit, xue2024follow, zhu2024multibooth, Sketchdeco, Magicanimate, Magicmix}. Reference-based colorization~\cite{Plug-and-play, meng2024anidoc, wang2024taming, locate, Cones, Cones2}, guided by a user-provided reference image, has also become popular, with existing methods using strategies such as segmented graphs, active learning, and attention networks. AnimeDiffusion~\cite{Animediffusion} and ColorizeDiffusion~\cite{Colorizediffusion} introduce a diffusion-based reference-based framework for anime face colorization. Paint-by-Example~\cite{Paintbyexample} and ObjectStitch~\cite{Objectstitch} leverage CLIP as their cross-modal encoder to extract instance-level visual-semantic embeddings, whereas AnyDoor~\cite{Anydoor} innovates by training on video sequences and adopting DINOv2~\cite{Dinov2} for spatial-temporal feature learning. Despite these advancements, all frameworks focus on generic object categories, falling short of fine-grained part-level alignment required for intricate design tasks~\cite{wang2024diffusion, wang2025magicscroll}.

\section{Preliminaries}

\subsection{Latent Diffusion Model}
As the core architecture of Stable Diffusion~\cite{Ldm}, Latent Diffusion Models (LDM) revolutionize text-to-image generation by executing diffusion-denoising processes in a compressed latent space rather than the raw pixel domain, enabling stable and efficient training. The pipeline begins with a Variational Autoencoder (VAE) projecting RGB images into low-dimensional latent codes, where semantic-guided diffusion sampling occurs under textual conditioning. Then, a UNet-based network incorporates self-attention and cross-attention mechanisms through UNet blocks to learn the reverse denoising process in the latent space. Cross-attention establishes bidirectional interactions between text embeddings and visual features, ensuring prompt semantics are continuously infused. The whole training objective of the UNet can be written as:
\begin{equation}
\mathcal{L}_{LDM} = \mathbb{E}_{t, \mathbf{z}, \epsilon} \left[ \left\| \epsilon - \epsilon_{\theta} \left( \sqrt{\alpha_t} \mathbf{z} + \sqrt{1 - \alpha_t} \epsilon, c, t \right) \right\|^2 \right],
\label{equ:ldm}
\end{equation}
where $z$ notes the latent embedding of the training sample. \(\epsilon_{\theta}\) and \(\epsilon\) represent predicted noise by the diffusion model and ground truth noise at corresponding timestep $t$, respectively. $c$ is the condition embedding involved in the generation, and the coefficient \(\alpha_{t}\) remains consistent with that employed in vanilla diffusion models.

\subsection{Reference Condition Injection}
In the current scenario, reference instances frequently suffer from background noise, redundancy, and semantic conflicts (e.g., similar instances coexist in the same reference image, or the background contains complex textures), which significantly impede the model's ability to learn colors effectively. To address this issue, when presented with $N$ user-provided reference instances along with the line art input, our model employs a dual-branch condition injection strategy. This strategy aims to achieve semantic similarity and structural alignment with the input.
First, we align multiple instances in a layer and input them into the reference net. Then, we encode with CLIP and apply reference attention mechanisms~\cite{Prompt2prompt, Masactrl, manganinja} to inject color and semantic info into the UNet. Also, the line art is injected into the UNet to enhance line-structure features. This lets the model learn global reference info evenly and avoid interference.
Second, for instance-aware image generation, we utilize instance images \(I_i\) embedding as latent control signals. Unlike other methods~\cite{Dreambooth, manganinja}, our paradigm supports the use of multiple instances, which enhances the model's generalizability. Zero-shot customization is tough, so we use pre-trained vision models to extract the target object identity. Previous studies used CLIP for target embedding~\cite{Omnibooth, Clip}. We use DINOv2 as the feature encoder~\cite{Dinov2} to get discriminative spatial identity. DINOv2, trained with patch-level objectives under random masking, has highly expressive features. Its output includes a \(26\times26\times1024\) spatial embedding \(s_i\) for patch-level features and a \(1024\) dimensional global embedding \(g_i\).

\begin{figure*}[htbp]
  \centering
  \includegraphics[width=\linewidth]{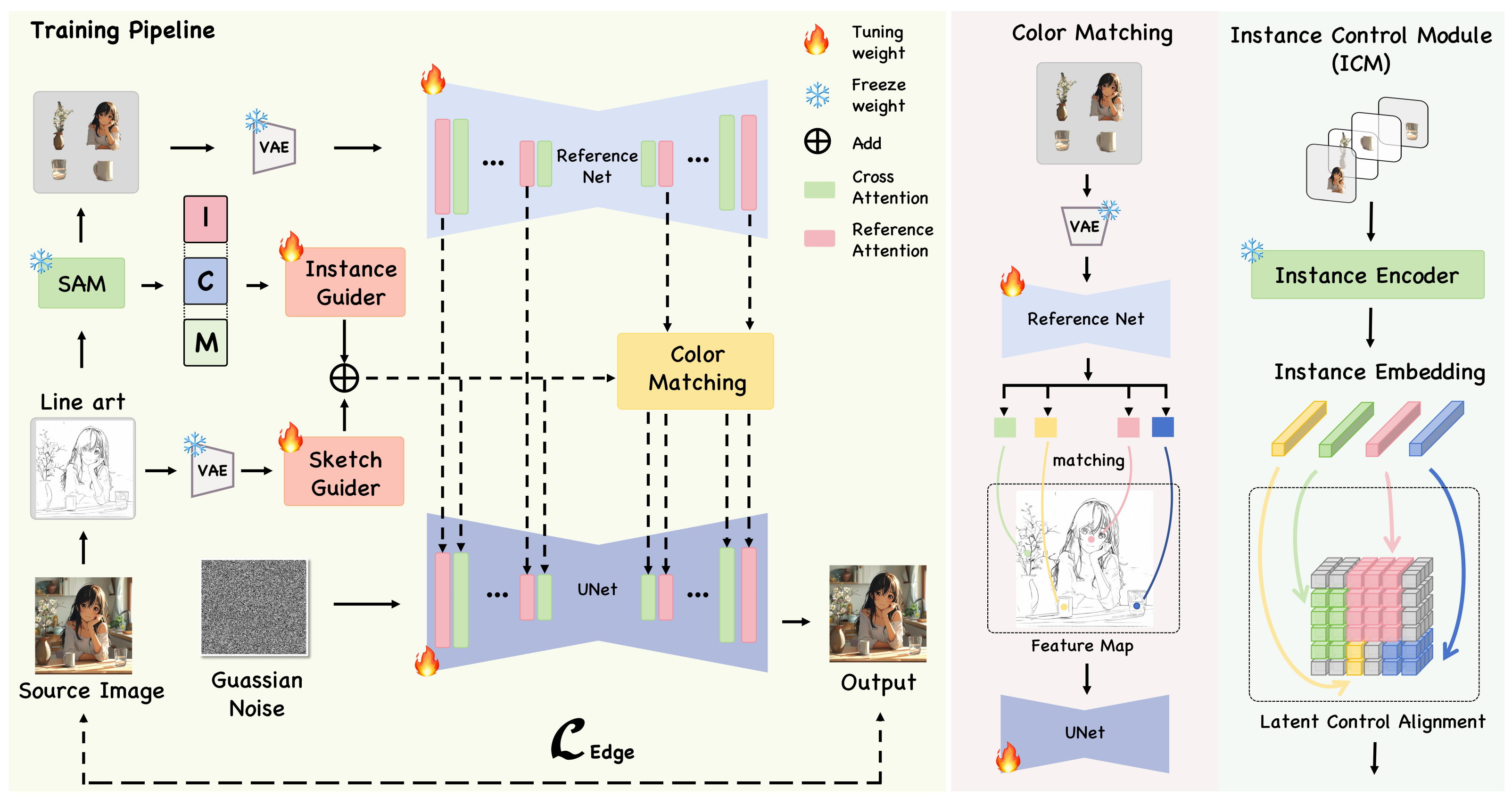}
  \caption{\textbf{Overview of the Follow-Your-Color pipeline.} We combine a dual-UNet framework with an instance control module (ICM). During training, we use multiple instances to set the overall color accurately. The color matching enables the model to better align the colors of the target image with those of the reference instances precisely. The edge loss helps the model pay more attention to the high-frequency areas and edges, resulting in a more accurate and vivid colorization for each instance. }
  \label{fig: pipeline}
\end{figure*}

\section{Method}
\noindent\textbf{Problem Definition.}
We first formulate the task of multi-instance sketch colorization with per-instance chromatic control, enabling precise mapping from multiple reference objects to corresponding line drawing instances via semantic correspondence. We can represent the problem in a more concise way using the following formula:
\begin{align}
    I_{out} = (S, R, M),
\end{align}
where \(I_{out}\) is the output image to be generated. The condition is composed of a line art image \(S\), a set of reference instances \(R = \{[I_{1},\cdots, I_{i}] \mid i = 1,\cdots, N\}\), and a set of instance masks \(M=\{[M_{1},\cdots, M_{i}] \mid i = 1,\cdots, N\}\). Each \(I_{i}\) in \(R\) corresponds to an instance, and each \(M_{i}\) in \(M\) is a binary mask indicating the reference spatial location of an instance.

\noindent\textbf{Architecture Design.} 
Owing to the high-fidelity detail demands in anime sketch colorization, the core challenge resides in designing encoders capable of spatially precise reference image parsing to extract pixel-level visual cues. Inspired by recent studies~\cite{animateanyone, anydressing, manganinja} which demonstrated the effectiveness of leveraging an additional UNet architecture for this purpose, we introduce a reference net following a similar design. Additionally, the sketch and instance guider are designed based on ControlNet~\cite{controlnet}, which provides an efficient condition for injecting line art and instance-level information into the generative process. Furthermore, we employ DINOv2~\cite{Dinov2} to encode images and train a Feed-Forward Network layer to extract features from the image embeddings. These extracted features are then integrated into the latent control signals.

\subsection{Self-Play Training Strategy}
\label{Two-stage Training}
We then introduce a two-stage, self-play training approach to solve the problem of a lack of multi-instance training data and gradually boost the model's performance to build an advanced model that can give high-quality coloring results.

\noindent\textbf{Single-Reference Colorization Training.} In the first stage, training starts by activating the reference net, UNet, and the sketch guider. In both stages, we use color matching with edge loss for training. For each anime sequence, we perform random frame sampling without replacement: one frame is designated as the style reference exemplar, while the other serves as the raw input sketch for colorization. We use the whole reference image as a conditional input and extract the line art from the original image. Given a reference and line art image pair, the model maps color semantics from the reference image to the line art by minimizing a carefully designed edge loss function, aiming to make the model learn basic color-semantic relationships, forming the basis for later multi-instance refinement.

\noindent\textbf{Multi-Instance Refinement.} In the second stage, we train the reference net, UNet, sketch guider, and instance guider to improve the model's ability to handle multiple instances and make local coloring more accurate. To precisely control each instance in the references, we use the Segment Anything Model (SAM)~\cite{Sam, Ram, Groundingdino} to extract instances separately and perform operations like random fusion, scaling, shuffling, and adding noise to instances in a layer as reference net input to enhance the model's per-instance semantic perception for accurate colorization. Then, the model encodes all instances, aligning them in latent space, and starts coloring each instance from the sketch, considering their features and overall relationships. This process results in more accurate colorization, especially in areas with high color and semantic requirements.

\subsection{Instance Control Module}
\begin{figure}
  \includegraphics[width=\linewidth]{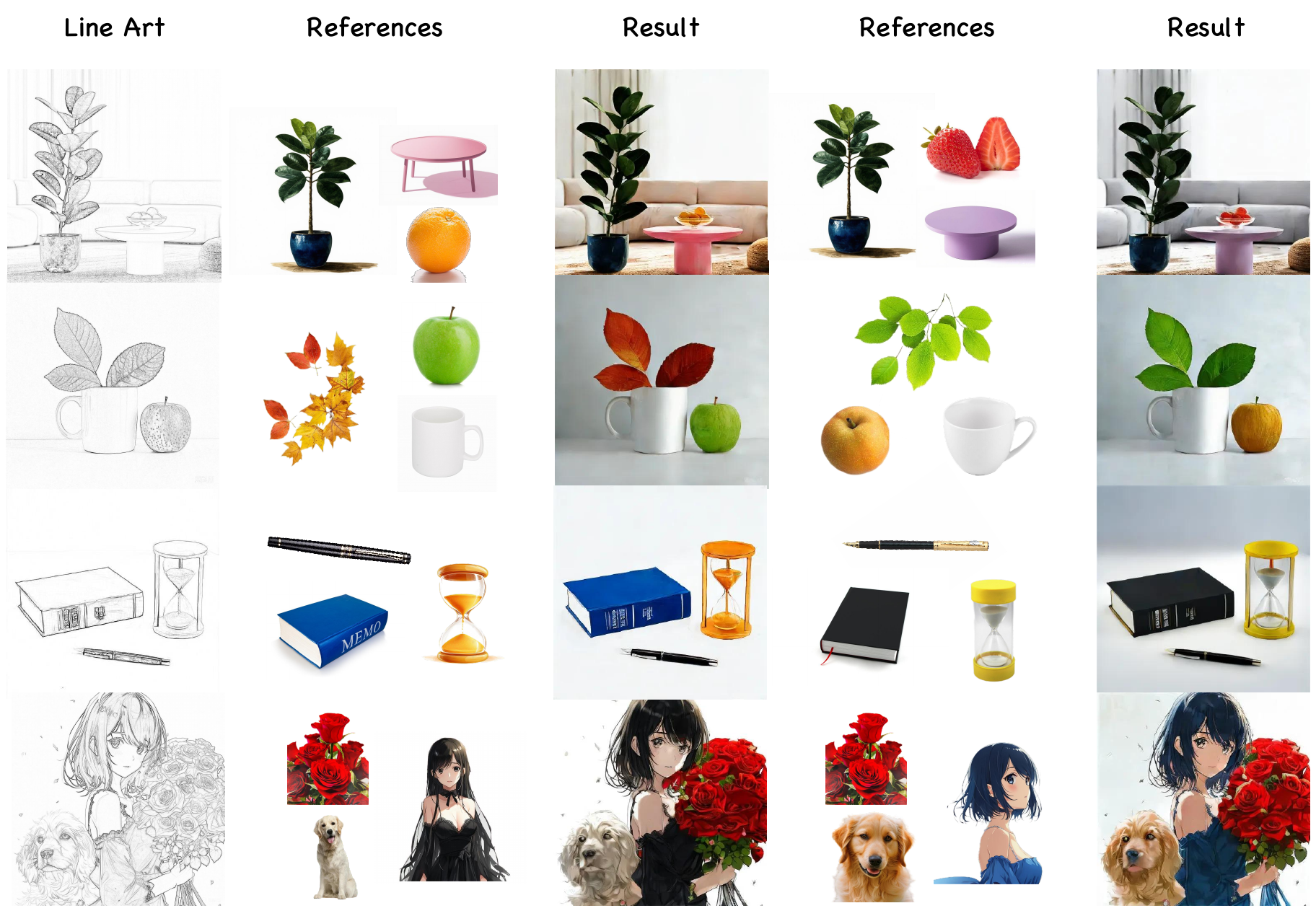}
  \captionof{figure}{\textbf{Instance Control Ability.} With the same line art and diverse reference instances, our method achieves precise instance-level control for varied colorization and all without needing extra guidance.}
  \label{control}
\end{figure}

Previous methods often resorted to directly imposing constraints within the reference attention mechanism for control purposes~\cite{Prompt2prompt, Masactrl, animateanyone}. Nevertheless, when faced with the task of handling multiple instances as conditional input, this conventional approach proved to be extremely arduous. To uphold semantic correspondence and safeguard details, we align image embeddings in the latent space, thereby empowering more precise and effective handling of the intricate input scenarios presented by multiple instances. Given target regions (instance masks), we align the bounding rectangle of the mask with the input DINO feature maps using box coordinates. Subsequently, we use the dense grids of the mask within this rectangle to interpolate features from the input DINO features. This process is similar to ROI alignment, but the resolution of the ROI is flexible and follows the size of the mask. Finally, to further incorporate global information, we randomly drop 10\% of the spatial embedding \(s_{i}\) and replace it with the DINO global embedding \(g_{i}\). As our target mask may have a different silhouette from the shape of the input image, injecting global embedding enhances the model's generalizability and adaptability across diverse image conditions. 
The final ROI feature is then warped into the original region of the latent control signal:
\begin{equation}
    l_{c}=\sum_{i = 1}^{n} Drop\_out(Interpolate(s_{i}, M_{i}, B_{i}), g_{i}),
\end{equation}
where $B_{i}$ is the bounding box of the target mask $M_{i}$. To establish an instance-aware colorization framework, we use a latent control signal, denoted as $l_c$, which is a latent feature with a size of $l_c\in\mathbb{R}^{C\times H'\times W'}$. As shown in Figure~\ref{control}, our model with an instance control module achieves precise instance-level control to produce varied colorization outcomes. Although the training data is derived from anime datasets, the references can incorporate real-life images, highlighting the model’s adaptability across different domains and flexible utilization of diverse reference sources.

\setcounter{table}{2}
\begin{table*}[h]
\caption{\textbf{Quantitative Comparisons Across Datasets}}
\label{crossdataset}
\centering
\resizebox{\textwidth}{!}{ 
\begin{tabular}{@{}lcc|cc|cc|cc@{}}
\toprule
\multirow{2}{*}{Domain Dataset} & 
\multicolumn{2}{c}{FID $\downarrow$} & 
\multicolumn{2}{c}{PSNR $\uparrow$} & 
\multicolumn{2}{c}{SSIM $\uparrow$} & 
\multicolumn{2}{c}{LPIPS $\downarrow$} \\
\cmidrule(lr){2-3} \cmidrule(lr){4-5} \cmidrule(lr){6-7} \cmidrule(lr){8-9}
& Animation & Hand-drawn & Animation & Hand-drawn & Animation & Hand-drawn & Animation & Hand-drawn \\
\midrule
RSIC(256x)          & 128.971 & 300.636 & 10.933 & 8.872 & 0.548 & 0.416 & 0.508 & 0.566 \\
SGA(256x)            & 124.732 & 306.881 & 10.021 & 7.253 & 0.521 & 0.414 & 0.454 & 0.533 \\
AnimeDiffusion(256x) & 154.784 & 305.021 & 11.262 & 10.207 & 0.416 & 0.426 & 0.557 & 0.549 \\
ColorizeDiffusion(512x) & 82.201 & 298.673 & 17.882 & 9.817 & 0.512 & 0.349 & 0.425 & 0.513 \\
MangaNinja(512x)     & 43.165  & 295.223  & 14.289 & 10.249 & 0.543 & 0.361 & 0.362 & 0.508 \\
\textbf{Ours(512x)}  & \textbf{28.953} & \textbf{251.301} & \textbf{23.751} & \textbf{10.551} & \textbf{0.783} & \textbf{0.434} & \textbf{0.201} & \textbf{0.431} \\
\bottomrule
\end{tabular}
}
\label{Quantitative result}
\end{table*}

\subsection{Structure-Content Supervision Enhancement} 
We present a comprehensive approach to enhance the performance of diffusion models in line art colorization. By introducing edge loss and color matching, we address crucial aspects of image quality, resulting in more visually appealing and perceptually accurate generated images. 

\noindent\textbf{Edge Loss.} During the diffusion model's training process, each pixel in an image contributes equally to the supervision process. The diffusion model's original training objective is pixel-level mean squared error (MSE) loss. However, simultaneously, we aim to make the image content more consistent with human perception of image quality. To enhance the supervision over the high-frequency area and improve the generation quality, we propose the edge loss, which consists of the perceptual loss and the re-weighted edge loss.
The formula for the perceptual loss is as follows:
\begin{equation}
\mathcal{L}_{\text{perceptual}}=\frac{1}{N}\sum_{i = 1}^{N}\left|\phi_{i}(z)-\phi_{i}(\hat{z})\right|_{2}^{2},
\end{equation}
where \(\hat{z}\) is the prediction latent embedding obtained by decoding $\epsilon_{\theta}$, \(N\) represents the number of feature layers used for calculation, and \(\phi_{i}\) is the feature-extraction function corresponding to the \(i\)-th layer of the pre-trained neural network. \(\lambda\) is a hyper-parameter that balances \(\mathcal{L}_{LDM}\) and the perceptual loss \(\mathcal{L}_{perceptual}\). Moreover, in scenes characterized by complex hierarchical structures and overlapping objects, the importance of object edges may differ significantly from that of the plain background pixels. We calculate the overall instances edge map, which effectively mitigates interference from the background as we focus solely on the edges of individual objects. Then, we apply structure foreground enhancement from SyntheOcc~\cite{Syntheocc} to the edge pixels. Since we perform edge detection in the latent space with the size of the input images, we regard the edge map as a loss weight map \(w\) to enhance structure supervision.  Finally, our total loss can be written as:
\begin{align}
\mathcal{L}&= w \cdot \mathcal{L}_{LDM} + \lambda\cdot\mathcal{L}_{perceptual}
\end{align}

\noindent\textbf{Color Matching.} We employ pre-trained diffusion models to identify corresponding points within real-world images. Our model, which has a UNet at its core, processes noisy images by cleaning them and extracting features crucial for establishing correspondences. Two steps utilize a diffusion feature map to match pixels between two images. 
\textbf{(1) Semantic Matching.} This is accomplished by determining the nearest neighbors and computing the similarity using cosine distance. To obtain pixel correspondences, we begin by extracting dense features from both images and matching them. Let $F_{i}$ be the feature map of an image $i$. For a pixel at position $p$, we obtain the feature vector $F(p)$ through bilinear interpolation. In terms of color matching, $C_{reference}$ denotes the color feature of the reference image, and $C_{source}$ denotes the color feature of the source image. We obtain $C_{source}$ and $C_{reference}$ by sampling features from the normalized source and target features. To quantify their similarity, we calculate the Euclidean distance ($D$) between $C_{reference}$ and the flattened $C_{source}$. Given two feature vectors $v_1 = C_{reference}$ and $v_2 = C_{source}$, the Euclidean distance is computed as follows:
\begin{equation}
D=\sqrt{\sum_{k = 1}^{f}(v_{1k} - v_{2k})^2},
\end{equation}
where $f$ represents the feature dimension. Subsequently, we find the nearest-neighbor indices.
\textbf{(2) Feature injection}. Given a pixel $p_1$ in $v_1$, we find the corresponding pixel $p_2$ in $v_2$ as follows:
\begin{equation}
p_2= \mathop{\arg\min}_{p} \ \
  d\left(F_1\left(p_1\right), F_2(p)\right),
\end{equation}
where $d$ is the cosine distance. This approach transfers color-related information between the two images.

\begin{figure*}[htbp]
  \centering
  \includegraphics[width=0.8\linewidth]{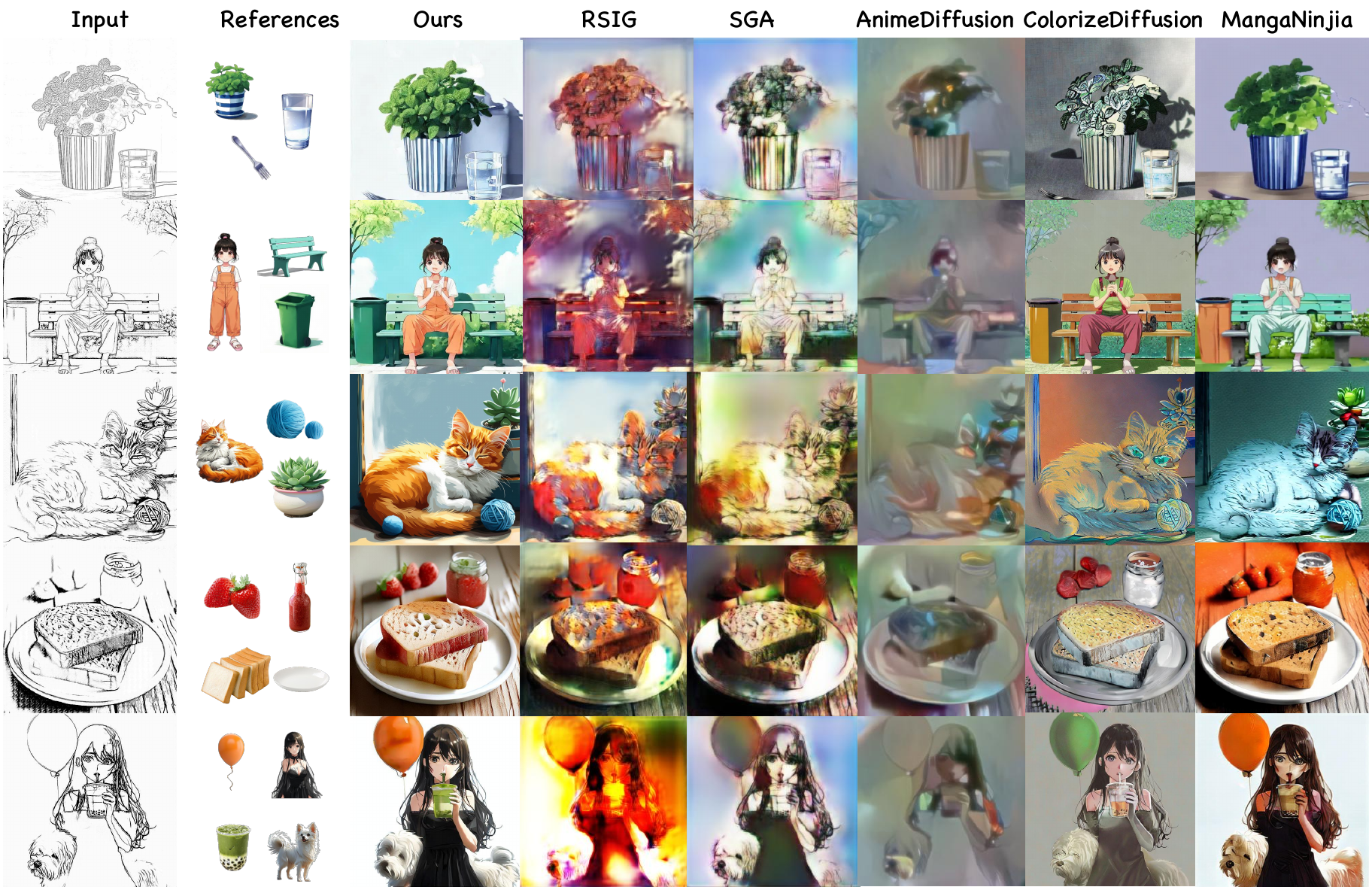}
  \caption{ \textbf{Qualitative comparisons with existing methods.} Given a line drawing and multiple reference instances, our method demonstrates far more precise colorization and higher-quality results compared to other methods, effectively maintaining line-drawing structure and reference instances' color consistency.}
  \label{comparsion}
\end{figure*}

\section{Experiments}
    
\subsection{Implementation Detail}
\noindent\textbf{Dataset.} 
The dataset employed in this study comprises two key components: the anime video dataset and the image dataset. The data were preprocessed from Sakuga~\cite{sakuga}, ATD-12K~\cite{Atd12k}, and manually collected from the internet. A total of 1,000 image pairs were selected from animations to form the ablation test set, and the remaining data constituted the training set. We then extracted their instances as references.

\noindent\textbf{Baseline design.} The backbone architecture for both the UNet and reference net is derived from Stable Diffusion 1.5~\cite{sd1.5}. The sketch guider and instance guider are initialized with pre-trained ControlNet weights. The instance encoder is initialized using DINOv2. 

\noindent\textbf{Hyper-parameters.} During training, we resized the height and width of the input image to 512 and kept the original aspect ratio. Our model undergoes 100,000 optimization iterations with a batch size of 1. The learning rate is set to \(1 \times 10^{-5} \). The training phase takes around 7 days using 2 NVIDIA A800 80G GPUs. We perform a random horizon flip and random brightness adjustment for each reference image as data augmentation to simulate multi-view conditions.

\subsection{Comparison}

\subsubsection{Qualitative Results} 
To achieve the same settings, we use a complete reference image as the reference condition for the comparative baseline methods, and our method employs instances extracted from the same reference image. We compare our approach with previous sketch colorization methods. For GAN-based approaches, we consider RSIC~\cite{RSIC} and SGA~\cite{Eliminatinggradient}. RSIC and SGA utilize GANs for colorization, each with a distinct architecture and training strategy. We observe that GAN-based methods tend to produce color incoordination. Moreover, when there are numerous color block regions in the image, color bleeding issues are likely to occur, where the color of one color block contaminates the adjacent color blocks. Regarding diffusion-based models, AnimeDiffusion~\cite{Animediffusion}, ColorizeDiffusion~\cite{Colorizediffusion}, and MangaNinja~\cite{manganinja} (without point guided) perform better in terms of color control. However, as shown in Figure~\ref{comparsion}, they still struggle to effectively transfer the colors from the reference instances to the line drawing sketch, especially for the colors of smaller details in the images. The root cause is that the models fail to fully learn the color correspondences between the original image and the reference instances.

\subsubsection{Quantitative Results} 
To comprehensively evaluate the colorization ability of our model, we quantitatively compared our method with the state-of-the-art colorization method on our test set. We constructed an \textbf{Animation Dataset} (1,570 pairs) from two anime films (\textit{Your Name} and \textit{Spirited Away}) and a \textbf{Hand-Drawn Dataset}~(100 pairs) collected from real-user sketches as a test set.  The results are presented in Table~\ref{Quantitative result}. We did not consider background noise when evaluating the experimental results. Due to the limited resolution of most previous work, all measurements are performed at a resolution of $512\times512$. We report four metrics of different methods. \textbf{FID} gauges visual similarity, with lower values indicating better quality. \textbf{PSNR} measures distortion, and higher values imply less distortion. \textbf{SSIM} assesses luminance, contrast, and structure similarity, with values closer to 1 showing better structural preservation. \textbf{LPIPS} measures perceptual similarity, and lower values mean closer resemblance to the reference. By comparing our model with these GAN-based and diffusion-based methods using the established metrics, our approach demonstrates a significant advantage over previous methods.
\begin{table}
\caption{\textbf{Quantitative results of ablation study.}}
\label{tab:freq}
\resizebox{\columnwidth}{!}{ 
\begin{tabular}{ccccc}
    \toprule
    ~ & FID $\downarrow$ & PSNR $\uparrow$ & SSIM $\uparrow$ & LPIPS $\downarrow$ \\
    \midrule
    All  & \textbf{62.531} & \textbf{22.587} & \textbf{0.806} & \textbf{0.203} \\
    w/o Edge Loss  & 83.042 & 20.842 & 0.755 & 0.217 \\
    w/o Instance Guider  & 91.852 & 16.467 & 0.718 & 0.308 \\
    w/o Color Matching  & 88.573 & 20.832 & 0.749 & 0.226 \\
    \bottomrule
    \label{ablation study}
\end{tabular}
}
\end{table}

\subsection{Ablation Study}

\begin{figure}[htbp]
  \centering
  \includegraphics[width=\linewidth]{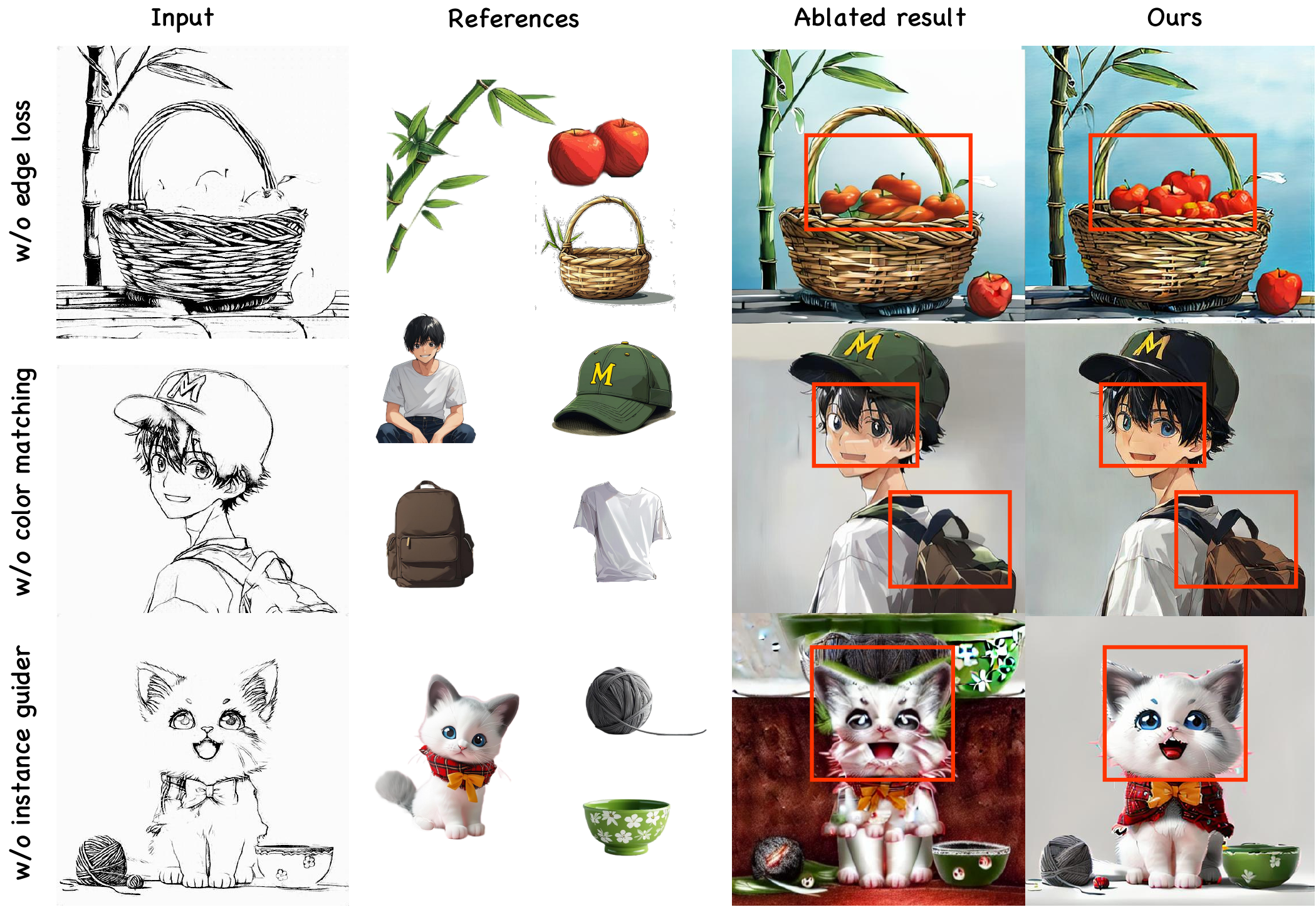}
  \caption{\textbf{Ablations on each component.} ``w/o edge loss" indicates without the edge loss, ``w/o color matching" indicates without the color matching, ``w/o instance guider" indicates without the instance guider.}
  \label{ablation}
\end{figure}

\noindent\textbf{Effectiveness of Edge Loss.} To evaluate edge loss, we replace its edge loss with the original diffusion loss. Our method fails to preserve structural edge information, resulting in mismatched colors on the apple's edges, as shown in the first row of Figure~\ref{ablation}. When we eliminate the overall structure edge loss, the model struggles to maintain subtle edge features. Numerical evaluations in Table~\ref{ablation study} confirm the importance of structure edge loss for preserving image integrity.

\noindent\textbf{Effectiveness of Color Matching.} Removing the color-matching component diminishes our model's ability to retain color information and details from input references. For instance, in the second row of Figure~\ref{ablation}, the colors of the bag and the character's facial details are inconsistent.

\noindent\textbf{Effectiveness of Instance Guider.} To illustrate the effectiveness of our instance guider, we remove it during the experiment. Visual results in Figure~\ref{ablation} show that without this module, our model struggles to transfer instance-level color information from references effectively. As can be seen in the third row, the lack of an instance guider hampers the learning of instance-level information, leading to significant color loss and noise in the final results when relying solely on the reference net.

\subsection{Limitations and Discussion}
Our method has many advantages and potential applications, but it also faces several limitations. We summarize them as follows: (1) \textbf{Flexible usage:} When stylizing the same sketch with diverse reference images, our method retains the character's identity. It adjusts details like lighting and background based on the reference styles. (2) \textbf{Semantic awareness:} For cartoon wallpapers or posters with multiple characters or objects, our approach uses multiple reference images to perform semantic-based colorization, ensuring each line art element gets colorized semantically. (3) \textbf{Multi-object and occlusions:} Single-subject line art can efficiently convert a sketch into a vivid, full-colored illustration, speeding up the production and enabling animators to test different color concepts quickly. However, in sketch images with many main objects or significant occlusions between them, the detailed inter-object colors may not be well maintained. (4) \textbf{More explorations:} The model can accurately color each element in multi-subject scenarios such as an anime battle scene or a forest scene while maintaining a harmonious overall color scheme.

\section{Conclusion}
This paper has presented \textit{Follow-Your-Color}, a diffusion-based sketch colorization framework. Specifically, we introduce a multi-instance approach, a two-stage, self-play training strategy, and pixel-level color matching with edge loss. Our experiments demonstrated that \textit{Follow-Your-Color} outperforms current methods in visual quality and style consistency, advancing the field of digital cartoon colorization. In future work, we plan to substantially augment the number of reference instances and enhance the model's capacity to maintain semantic and color consistency.

\section{Acknowledagment}
This work was partially supported by the Guangdong Provincial Key Lab of Integrated Communication, Sensing and Computation for Ubiquitous Internet of Things \#2023B1212010007. The Research Grant Council of the Hong Kong Special Administrative Region under grant number 16203122.

{
    \small
    \bibliographystyle{ieeenat_fullname}
    \bibliography{main}

\begin{thebibliography}{104}
\providecommand{\natexlab}[1]{#1}
\providecommand{\url}[1]{\texttt{#1}}
\expandafter\ifx\csname urlstyle\endcsname\relax
  \providecommand{\doi}[1]{doi: #1}\else
  \providecommand{\doi}{doi: \begingroup \urlstyle{rm}\Url}\fi

\bibitem[Bai et~al.(2022)Bai, Dong, Chai, Wang, Xu, and Yuan]{example3}
Yunpeng Bai, Chao Dong, Zenghao Chai, Andong Wang, Zhengzhuo Xu, and Chun Yuan.
\newblock {Semantic-Sparse Colorization Network for Deep Exemplar-Based Colorization}.
\newblock In \emph{European Conference on Computer Vision}, pages 505--521. Springer, 2022.

\bibitem[Bay et~al.(2006)Bay, Tuytelaars, and Van~Gool]{surf}
Herbert Bay, Tinne Tuytelaars, and Luc Van~Gool.
\newblock {SURF: Speeded Up Robust Features}.
\newblock In \emph{Computer Vision--ECCV 2006: 9th European Conference on Computer Vision, Graz, Austria, May 7-13, 2006. Proceedings, Part I 9}, pages 404--417. Springer, 2006.

\bibitem[Brooks et~al.(2023)Brooks, Holynski, and Efros]{Instructpix2pix}
Tim Brooks, Aleksander Holynski, and Alexei~A Efros.
\newblock {InstructPix2Pix: Learning to Follow Image Editing Instructions}.
\newblock In \emph{Proceedings of the IEEE/CVF Conference on Computer Vision and Pattern Recognition}, pages 18392--18402, 2023.

\bibitem[Cao et~al.(2023{\natexlab{a}})Cao, Wang, Qi, Shan, Qie, and Zheng]{Masactrl}
Mingdeng Cao, Xintao Wang, Zhongang Qi, Ying Shan, Xiaohu Qie, and Yinqiang Zheng.
\newblock {Masactrl: Tuning-Free Mutual Self-Attention Control for Consistent Image Synthesis and Editing}.
\newblock In \emph{Proceedings of the IEEE/CVF International Conference on Computer Vision}, pages 22560--22570, 2023{\natexlab{a}}.

\bibitem[Cao et~al.(2021)Cao, Mo, and Gao]{Regionmap}
Ruizhi Cao, Haoran Mo, and Chengying Gao.
\newblock {Line Art Colorization Based on Explicit Region Segmentation}.
\newblock In \emph{Computer Graphics Forum}, pages 1--10. Wiley Online Library, 2021.

\bibitem[Cao et~al.(2023{\natexlab{b}})Cao, Meng, Mok, Liu, Lee, and Li]{Animediffusion}
Yu Cao, Xiangqiao Meng, PY Mok, Xueting Liu, Tong-Yee Lee, and Ping Li.
\newblock {AnimeDiffusion: Anime Face Line Drawing Colorization via Diffusion Models}.
\newblock \emph{arXiv preprint arXiv:2303.11137}, 2023{\natexlab{b}}.

\bibitem[Carrillo et~al.(2023)Carrillo, Cl{\'e}ment, Bugeau, and Simo-Serra]{Diffusart}
Hernan Carrillo, Micha{\"e}l Cl{\'e}ment, Aur{\'e}lie Bugeau, and Edgar Simo-Serra.
\newblock {Diffusart: Enhancing Line Art Colorization with Conditional Diffusion Models}.
\newblock In \emph{Proceedings of the IEEE/CVF Conference on Computer Vision and Pattern Recognition}, pages 3486--3490, 2023.

\bibitem[Casey et~al.(2021)Casey, P{\'e}rez, and Li]{Animationtransformer}
Evan Casey, V{\'\i}ctor P{\'e}rez, and Zhuoru Li.
\newblock The animation transformer: Visual correspondence via segment matching.
\newblock In \emph{Proceedings of the IEEE/CVF International Conference on Computer Vision}, pages 11323--11332, 2021.

\bibitem[Chen et~al.(2024{\natexlab{a}})Chen, Ma, Wang, Yuan, Zhao, Tian, Wang, Min, Chen, and Liu]{chen2024follow}
Qihua Chen, Yue Ma, Hongfa Wang, Junkun Yuan, Wenzhe Zhao, Qi Tian, Hongmei Wang, Shaobo Min, Qifeng Chen, and Wei Liu.
\newblock {Follow-Your-Canvas: Higher-Resolution Video Outpainting with Extensive Content Generation}.
\newblock \emph{arXiv preprint arXiv:2409.01055}, 2024{\natexlab{a}}.

\bibitem[Chen et~al.(2024{\natexlab{b}})Chen, Huang, Liu, Shen, Zhao, and Zhao]{Anydoor}
Xi Chen, Lianghua Huang, Yu Liu, Yujun Shen, Deli Zhao, and Hengshuang Zhao.
\newblock {Anydoor: Zero-Shot Object-Level Image Customization}.
\newblock In \emph{Proceedings of the IEEE/CVF Conference on Computer Vision and Pattern Recognition}, pages 6593--6602, 2024{\natexlab{b}}.

\bibitem[Chia et~al.(2011)Chia, Zhuo, Gupta, Tai, Cho, Tan, and Lin]{example5}
Alex Yong-Sang Chia, Shaojie Zhuo, Raj~Kumar Gupta, Yu-Wing Tai, Siu-Yeung Cho, Ping Tan, and Stephen Lin.
\newblock {Semantic Colorization With Internet Images}.
\newblock \emph{ACM Transactions on Graphics (TOG)}, 30\penalty0 (6):\penalty0 1--8, 2011.

\bibitem[Cho et~al.(2021)Cho, Hong, Jeon, Lee, Sohn, and Kim]{cats}
Seokju Cho, Sunghwan Hong, Sangryul Jeon, Yunsung Lee, Kwanghoon Sohn, and Seungryong Kim.
\newblock {CATs: Cost Aggregation Transformers for Visual Correspondence}.
\newblock \emph{Advances in Neural Information Processing Systems}, 34:\penalty0 9011--9023, 2021.

\bibitem[Dou et~al.(2021)Dou, Wang, Li, Wang, Li, and Liu]{Dualspace}
Zhi Dou, Ning Wang, Baopu Li, Zhihui Wang, Haojie Li, and Bin Liu.
\newblock {Dual Color Space Guided Sketch Colorization}.
\newblock \emph{IEEE Transactions on Image Processing}, 30:\penalty0 7292--7304, 2021.

\bibitem[Feng et~al.(2024)Feng, Ma, Wang, Qi, Chen, Chen, and Wang]{feng2024dit4edit}
Kunyu Feng, Yue Ma, Bingyuan Wang, Chenyang Qi, Haozhe Chen, Qifeng Chen, and Zeyu Wang.
\newblock {Dit4Edit: Diffusion Transformer for Image Editing}.
\newblock \emph{arXiv preprint arXiv:2411.03286}, 2024.

\bibitem[Gal et~al.(2022)Gal, Alaluf, Atzmon, Patashnik, Bermano, Chechik, and Cohen-Or]{Textinversion}
Rinon Gal, Yuval Alaluf, Yuval Atzmon, Or Patashnik, Amit~H Bermano, Gal Chechik, and Daniel Cohen-Or.
\newblock {An Image is Worth One Word: Personalizing Text-To-Image Generation Using Textual Inversion}.
\newblock \emph{arXiv preprint arXiv:2208.01618}, 2022.

\bibitem[Goodfellow et~al.(2014)Goodfellow, Pouget-Abadie, Mirza, Xu, Warde-Farley, Ozair, Courville, and Bengio]{Gan}
Ian Goodfellow, Jean Pouget-Abadie, Mehdi Mirza, Bing Xu, David Warde-Farley, Sherjil Ozair, Aaron Courville, and Yoshua Bengio.
\newblock {Generative Adversarial Nets}.
\newblock \emph{Advances in neural information processing systems}, 27, 2014.

\bibitem[Guadarrama et~al.(2017)Guadarrama, Dahl, Bieber, Norouzi, Shlens, and Murphy]{Pixcolor}
Sergio Guadarrama, Ryan Dahl, David Bieber, Mohammad Norouzi, Jonathon Shlens, and Kevin Murphy.
\newblock {PixColor: Pixel Recursive Colorization}.
\newblock \emph{arXiv preprint arXiv:1705.07208}, 2017.

\bibitem[Hertz et~al.(2022)Hertz, Mokady, Tenenbaum, Aberman, Pritch, and Cohen-Or]{Prompt2prompt}
Amir Hertz, Ron Mokady, Jay Tenenbaum, Kfir Aberman, Yael Pritch, and Daniel Cohen-Or.
\newblock {Prompt-To-Prompt Image Editing with Cross Attention Control}.
\newblock \emph{arXiv preprint arXiv:2208.01626}, 2022.

\bibitem[Hu(2024)]{animateanyone}
Li Hu.
\newblock {Animate Anyone: Consistent and Controllable Image-To-Video Synthesis for Character Animation}.
\newblock In \emph{Proceedings of the IEEE/CVF Conference on Computer Vision and Pattern Recognition}, pages 8153--8163, 2024.

\bibitem[Huang et~al.(2020)Huang, Qiu, Wang, and Li]{example4}
Yifei Huang, Sheng Qiu, Changbo Wang, and Chenhui Li.
\newblock {Learning Representations for High-Dynamic-Range Image Color Transfer in a Self-Supervised Way}.
\newblock \emph{IEEE Transactions on Multimedia}, 23:\penalty0 176--188, 2020.

\bibitem[Huang et~al.(2022)Huang, Zhao, and Liao]{Unicolor}
Zhitong Huang, Nanxuan Zhao, and Jing Liao.
\newblock {UniColor: A Unified Framework for Multi-Modal Colorization With Transformer}.
\newblock \emph{ACM Transactions on Graphics (TOG)}, 41\penalty0 (6):\penalty0 1--16, 2022.

\bibitem[Iizuka et~al.(2016)Iizuka, Simo-Serra, and Ishikawa]{classification}
Satoshi Iizuka, Edgar Simo-Serra, and Hiroshi Ishikawa.
\newblock {Let There Be Color! Joint End-To-End Learning of Global and Local Image Priors for Automatic Image Colorization With Simultaneous Classification}.
\newblock \emph{ACM Transactions on Graphics (TOG)}, 35\penalty0 (4):\penalty0 1--11, 2016.

\bibitem[Ji et~al.(2022)Ji, Jiang, Luo, Tao, Chu, Xie, Wang, and Tai]{Colorformer}
Xiaozhong Ji, Boyuan Jiang, Donghao Luo, Guangpin Tao, Wenqing Chu, Zhifeng Xie, Chengjie Wang, and Ying Tai.
\newblock {Colorformer: Image Colorization via Color Memory Assisted Hybrid-Attention Transformer}.
\newblock In \emph{European Conference on Computer Vision}, pages 20--36. Springer, 2022.

\bibitem[Karaev et~al.(2024)Karaev, Makarov, Wang, Neverova, Vedaldi, and Rupprecht]{cotracker3}
Nikita Karaev, Iurii Makarov, Jianyuan Wang, Natalia Neverova, Andrea Vedaldi, and Christian Rupprecht.
\newblock {COTRACKER3: Simpler and Better Point Tracking by Pseudo-Labelling Real Videos}.
\newblock \emph{arXiv preprint arXiv:2410.11831}, 2024.

\bibitem[Karras(2019)]{StyleGan}
Tero Karras.
\newblock {A Style-Based Generator Architecture for Generative Adversarial Networks}.
\newblock \emph{arXiv preprint arXiv:1812.04948}, 2019.

\bibitem[Karras et~al.(2020)Karras, Laine, Aittala, Hellsten, Lehtinen, and Aila]{Stylegan2}
Tero Karras, Samuli Laine, Miika Aittala, Janne Hellsten, Jaakko Lehtinen, and Timo Aila.
\newblock {Analyzing and Improving the Image Quality of Stylegan}.
\newblock In \emph{Proceedings of the IEEE/CVF conference on computer vision and pattern recognition}, pages 8110--8119, 2020.

\bibitem[Kim et~al.(2019)Kim, Jhoo, Park, and Yoo]{Tag2pix}
Hyunsu Kim, Ho~Young Jhoo, Eunhyeok Park, and Sungjoo Yoo.
\newblock {Tag2pix: Line Art Colorization Using Text Tag With Secat and Changing Loss}.
\newblock In \emph{Proceedings of the IEEE/CVF International Conference on Computer Vision}, pages 9056--9065, 2019.

\bibitem[Kim et~al.(2022)Kim, Min, and Cho]{transformatcher}
Seungwook Kim, Juhong Min, and Minsu Cho.
\newblock {TransforMatcher: Match-To-Match Attention for Semantic Correspondence}.
\newblock In \emph{Proceedings of the IEEE/CVF conference on computer vision and pattern recognition}, pages 8697--8707, 2022.

\bibitem[Kirillov et~al.(2023)Kirillov, Mintun, Ravi, Mao, Rolland, Gustafson, Xiao, Whitehead, Berg, Lo, et~al.]{Sam}
Alexander Kirillov, Eric Mintun, Nikhila Ravi, Hanzi Mao, Chloe Rolland, Laura Gustafson, Tete Xiao, Spencer Whitehead, Alexander~C Berg, Wan-Yen Lo, et~al.
\newblock {Segment Anything}.
\newblock In \emph{Proceedings of the IEEE/CVF International Conference on Computer Vision}, pages 4015--4026, 2023.

\bibitem[Lee et~al.(2020)Lee, Kim, Lee, Kim, Chang, and Choo]{RSIC}
Junsoo Lee, Eungyeup Kim, Yunsung Lee, Dongjun Kim, Jaehyuk Chang, and Jaegul Choo.
\newblock {Reference-Based Sketch Image Colorization Using Augmented-Self Reference and Dense Semantic Correspondence}.
\newblock In \emph{Proceedings of the IEEE/CVF conference on computer vision and pattern recognition}, pages 5801--5810, 2020.

\bibitem[Lee et~al.(2021)Lee, DeGol, Fragoso, and Sinha]{patchmatch}
Jae~Yong Lee, Joseph DeGol, Victor Fragoso, and Sudipta~N Sinha.
\newblock {Patchmatch-Based Neighborhood Consensus for Semantic Correspondence}.
\newblock In \emph{Proceedings of the IEEE/CVF Conference on Computer Vision and Pattern Recognition}, pages 13153--13163, 2021.

\bibitem[Lei and Chen(2019)]{Videocolor}
Chenyang Lei and Qifeng Chen.
\newblock {Fully Automatic Video Colorization With Self-Regularization and Diversity}.
\newblock In \emph{Proceedings of the IEEE/CVF conference on computer vision and pattern recognition}, pages 3753--3761, 2019.

\bibitem[Li et~al.(2024{\natexlab{a}})Li, Qiu, Cai, Yan, Lian, Liu, and Chen]{Syntheocc}
Leheng Li, Weichao Qiu, Yingjie Cai, Xu Yan, Qing Lian, Bingbing Liu, and Ying-Cong Chen.
\newblock {SyntheOcc: Synthesize Geometric-Controlled Street View Images through 3D Semantic MPIs}.
\newblock \emph{arXiv preprint arXiv:2410.00337}, 2024{\natexlab{a}}.

\bibitem[Li et~al.(2024{\natexlab{b}})Li, Qiu, Yan, He, Zhou, Cai, Lian, Liu, and Chen]{Omnibooth}
Leheng Li, Weichao Qiu, Xu Yan, Jing He, Kaiqiang Zhou, Yingjie Cai, Qing Lian, Bingbing Liu, and Ying-Cong Chen.
\newblock {OmniBooth: Learning Latent Control for Image Synthesis with Multi-Modal Instruction}.
\newblock \emph{arXiv preprint arXiv:2410.04932}, 2024{\natexlab{b}}.

\bibitem[Li et~al.(2024{\natexlab{c}})Li, Sun, Zhang, Ye, Liao, Feng, Zhao, and He]{anydressing}
Xinghui Li, Qichao Sun, Pengze Zhang, Fulong Ye, Zhichao Liao, Wanquan Feng, Songtao Zhao, and Qian He.
\newblock {Anydressing: Customizable Multi-Garment Virtual Dressing via Latent Diffusion Models}.
\newblock \emph{arXiv preprint arXiv:2412.04146}, 2024{\natexlab{c}}.

\bibitem[Li et~al.(2022)Li, Geng, Kang, Chen, and Yang]{Eliminatinggradient}
Zekun Li, Zhengyang Geng, Zhao Kang, Wenyu Chen, and Yibo Yang.
\newblock {Eliminating Gradient Conflict in Reference-Based Line-Art Colorization}.
\newblock In \emph{European Conference on Computer Vision}, pages 579--596. Springer, 2022.

\bibitem[Liew et~al.(2022)Liew, Yan, Zhou, and Feng]{Magicmix}
Jun~Hao Liew, Hanshu Yan, Daquan Zhou, and Jiashi Feng.
\newblock {MagicMix: Semantic Mixing with Diffusion Models}.
\newblock \emph{arXiv preprint arXiv:2210.16056}, 2022.

\bibitem[Lindenberger et~al.(2023)Lindenberger, Sarlin, and Pollefeys]{lightglue}
Philipp Lindenberger, Paul-Edouard Sarlin, and Marc Pollefeys.
\newblock {Lightglue: Local Feature Matching at Light Speed}.
\newblock In \emph{Proceedings of the IEEE/CVF International Conference on Computer Vision}, pages 17627--17638, 2023.

\bibitem[Liu et~al.(2025{\natexlab{a}})Liu, Zeng, Ren, Li, Zhang, Yang, Jiang, Li, Yang, Su, et~al.]{Groundingdino}
Shilong Liu, Zhaoyang Zeng, Tianhe Ren, Feng Li, Hao Zhang, Jie Yang, Qing Jiang, Chunyuan Li, Jianwei Yang, Hang Su, et~al.
\newblock {Grounding Dino: Marrying Dino With Grounded Pre-training for Open-Set Object Detection}.
\newblock In \emph{European Conference on Computer Vision}, pages 38--55. Springer, 2025{\natexlab{a}}.

\bibitem[Liu et~al.(2022)Liu, Wu, Li, Li, and Wu]{Reference1}
Xueting Liu, Wenliang Wu, Chengze Li, Yifan Li, and Huisi Wu.
\newblock {Reference-Guided Structure-Aware Deep Sketch Colorization for Cartoons}.
\newblock \emph{Computational Visual Media}, 8:\penalty0 135--148, 2022.

\bibitem[Liu et~al.(2018)Liu, Qin, Wan, and Luo]{Auto-painter}
Yifan Liu, Zengchang Qin, Tao Wan, and Zhenbo Luo.
\newblock {Auto-Painter: Cartoon Image Generation from Sketch by Using Conditional Wasserstein Generative Adversarial Networks}.
\newblock \emph{Neurocomputing}, 311:\penalty0 78--87, 2018.

\bibitem[Liu et~al.(2023{\natexlab{a}})Liu, Feng, Zhu, Zhang, Zheng, Liu, Zhao, Zhou, and Cao]{Cones}
Zhiheng Liu, Ruili Feng, Kai Zhu, Yifei Zhang, Kecheng Zheng, Yu Liu, Deli Zhao, Jingren Zhou, and Yang Cao.
\newblock {Cones: Concept Neurons in Diffusion Models for Customized Generation}.
\newblock \emph{arXiv preprint arXiv:2303.05125}, 2023{\natexlab{a}}.

\bibitem[Liu et~al.(2023{\natexlab{b}})Liu, Zhang, Shen, Zheng, Zhu, Feng, Liu, Zhao, Zhou, and Cao]{Cones2}
Zhiheng Liu, Yifei Zhang, Yujun Shen, Kecheng Zheng, Kai Zhu, Ruili Feng, Yu Liu, Deli Zhao, Jingren Zhou, and Yang Cao.
\newblock {Cones 2: Customizable Image Synthesis With Multiple Subjects}.
\newblock In \emph{Proceedings of the 37th International Conference on Neural Information Processing Systems}, pages 57500--57519, 2023{\natexlab{b}}.

\bibitem[Liu et~al.(2025{\natexlab{b}})Liu, Cheng, Chen, Xiao, Ouyang, Zhu, Liu, Shen, Chen, and Luo]{manganinja}
Zhiheng Liu, Ka~Leong Cheng, Xi Chen, Jie Xiao, Hao Ouyang, Kai Zhu, Yu Liu, Yujun Shen, Qifeng Chen, and Ping Luo.
\newblock {MangaNinja: Line Art Colorization with Precise Reference Following}.
\newblock \emph{arXiv preprint arXiv:2501.08332}, 2025{\natexlab{b}}.

\bibitem[Lowe(2004)]{scale-invariant}
David~G Lowe.
\newblock {Distinctive Image Features from Scale-Invariant Keypoints}.
\newblock \emph{International Journal of Computer Vision}, 60:\penalty0 91--110, 2004.

\bibitem[Lyu et~al.(2019)Lyu, Sha, Qin, Yan, Xie, and Wang]{Ddpm}
He Lyu, Ningyu Sha, Shuyang Qin, Ming Yan, Yuying Xie, and Rongrong Wang.
\newblock {Advances in Neural Information Processing Systems}.
\newblock \emph{Advances in neural information processing systems}, 32, 2019.

\bibitem[Ma et~al.(2022)Ma, Wang, Wu, Lyu, Chen, Li, and Qiao]{ma2022visual}
Yue Ma, Yali Wang, Yue Wu, Ziyu Lyu, Siran Chen, Xiu Li, and Yu Qiao.
\newblock {Visual Knowledge Graph for Human Action Reasoning in Videos}.
\newblock In \emph{Proceedings of the 30th ACM International Conference on Multimedia}, pages 4132--4141, 2022.

\bibitem[Ma et~al.(2023)Ma, Cun, He, Qi, Wang, Shan, Li, and Chen]{ma2023magicstick}
Yue Ma, Xiaodong Cun, Yingqing He, Chenyang Qi, Xintao Wang, Ying Shan, Xiu Li, and Qifeng Chen.
\newblock {MagicStick: Controllable Video Editing via Control Handle Transformations}.
\newblock \emph{arXiv preprint arXiv:2312.03047}, 2023.

\bibitem[Ma et~al.(2024{\natexlab{a}})Ma, He, Cun, Wang, Chen, Li, and Chen]{ma2024followyourpose}
Yue Ma, Yingqing He, Xiaodong Cun, Xintao Wang, Siran Chen, Xiu Li, and Qifeng Chen.
\newblock {Follow Your Pose: Pose-Guided Text-To-Video Generation Using Pose-Free Videos}.
\newblock In \emph{Proceedings of the AAAI Conference on Artificial Intelligence}, pages 4117--4125, 2024{\natexlab{a}}.

\bibitem[Ma et~al.(2024{\natexlab{b}})Ma, He, Wang, Wang, Qi, Cai, Li, Li, Shum, Liu, et~al.]{ma2024followyourclick}
Yue Ma, Yingqing He, Hongfa Wang, Andong Wang, Chenyang Qi, Chengfei Cai, Xiu Li, Zhifeng Li, Heung-Yeung Shum, Wei Liu, et~al.
\newblock {Follow-Your-Click: Open-domain Regional Image Animation via Short Prompts}.
\newblock \emph{arXiv preprint arXiv:2403.08268}, 2024{\natexlab{b}}.

\bibitem[Ma et~al.(2024{\natexlab{c}})Ma, Liu, Wang, Pan, He, Yuan, Zeng, Cai, Shum, Liu, et~al.]{ma2024followyouremoji}
Yue Ma, Hongyu Liu, Hongfa Wang, Heng Pan, Yingqing He, Junkun Yuan, Ailing Zeng, Chengfei Cai, Heung-Yeung Shum, Wei Liu, et~al.
\newblock {Follow-Your-Emoji: Fine-Controllable and Expressive Freestyle Portrait Animation}.
\newblock \emph{arXiv preprint arXiv:2406.01900}, 2024{\natexlab{c}}.

\bibitem[Maejima et~al.(2019)Maejima, Kubo, Funatomi, Yotsukura, Nakamura, and Mukaigawa]{Graphmatching}
Akinobu Maejima, Hiroyuki Kubo, Takuya Funatomi, Tatsuo Yotsukura, Satoshi Nakamura, and Yasuhiro Mukaigawa.
\newblock {Graph Matching Based Anime Colorization With Multiple References}.
\newblock In \emph{ACM SIGGRAPH 2019 Posters}, pages 1--2, 2019.

\bibitem[Meng et~al.(2024)Meng, Ouyang, Wang, Wang, Wang, Cheng, Liu, Shen, and Qu]{meng2024anidoc}
Yihao Meng, Hao Ouyang, Hanlin Wang, Qiuyu Wang, Wen Wang, Ka~Leong Cheng, Zhiheng Liu, Yujun Shen, and Huamin Qu.
\newblock {AniDoc: Animation Creation Made Easier}.
\newblock \emph{arXiv preprint arXiv:2412.14173}, 2024.

\bibitem[Oquab et~al.(2023)Oquab, Darcet, Moutakanni, Vo, Szafraniec, Khalidov, Fernandez, Haziza, Massa, El-Nouby, et~al.]{Dinov2}
Maxime Oquab, Timoth{\'e}e Darcet, Th{\'e}o Moutakanni, Huy Vo, Marc Szafraniec, Vasil Khalidov, Pierre Fernandez, Daniel Haziza, Francisco Massa, Alaaeldin El-Nouby, et~al.
\newblock {DINOV2: Learning Robust Visual Features without Supervision}.
\newblock \emph{arXiv preprint arXiv:2304.07193}, 2023.

\bibitem[Pan et~al.(2024{\natexlab{a}})Pan, Mao, Jiang, Han, and Zhang]{locate}
Yulin Pan, Chaojie Mao, Zeyinzi Jiang, Zhen Han, and Jingfeng Zhang.
\newblock {Locate, Assign, Refine: Taming Customized Image Inpainting with Text-Subject Guidance}.
\newblock \emph{arXiv preprint arXiv:2403.19534}, 2024{\natexlab{a}}.

\bibitem[Pan et~al.(2024{\natexlab{b}})Pan, Zhu, and Mu]{sakuga}
Zhenglin Pan, Yu Zhu, and Yuxuan Mu.
\newblock {Sakuga-42M Dataset: Scaling Up Cartoon Research}.
\newblock \emph{arXiv preprint arXiv:2405.07425}, 2024{\natexlab{b}}.

\bibitem[Peebles et~al.(2022)Peebles, Zhu, Zhang, Torralba, Efros, and Shechtman]{Gan-supervised}
William Peebles, Jun-Yan Zhu, Richard Zhang, Antonio Torralba, Alexei~A Efros, and Eli Shechtman.
\newblock {Gan-Supervised Dense Visual Alignment}.
\newblock In \emph{Proceedings of the IEEE/CVF Conference on Computer Vision and Pattern Recognition}, pages 13470--13481, 2022.

\bibitem[Qu et~al.(2006)Qu, Wong, and Heng]{scrible2}
Yingge Qu, Tien-Tsin Wong, and Pheng-Ann Heng.
\newblock {Manga Colorization}.
\newblock \emph{ACM Transactions on Graphics (TOG)}, 25\penalty0 (3):\penalty0 1214--1220, 2006.

\bibitem[Radford et~al.(2021)Radford, Kim, Hallacy, Ramesh, Goh, Agarwal, Sastry, Askell, Mishkin, Clark, et~al.]{Clip}
Alec Radford, Jong~Wook Kim, Chris Hallacy, Aditya Ramesh, Gabriel Goh, Sandhini Agarwal, Girish Sastry, Amanda Askell, Pamela Mishkin, Jack Clark, et~al.
\newblock {Learning Transferable Visual Models from Natural Language Supervision}.
\newblock In \emph{International conference on machine learning}, pages 8748--8763. PMLR, 2021.

\bibitem[Rombach et~al.(2022{\natexlab{a}})Rombach, Blattmann, Lorenz, Esser, and Ommer]{Ldm}
Robin Rombach, Andreas Blattmann, Dominik Lorenz, Patrick Esser, and Bj{\"o}rn Ommer.
\newblock {High-Resolution Image Synthesis with Latent Diffusion Models}.
\newblock In \emph{Proceedings of the IEEE/CVF conference on computer vision and pattern recognition}, pages 10684--10695, 2022{\natexlab{a}}.

\bibitem[Rombach et~al.(2022{\natexlab{b}})Rombach, Blattmann, Lorenz, Esser, and Ommer]{sd1.5}
Robin Rombach, Andreas Blattmann, Dominik Lorenz, Patrick Esser, and Bj\"orn Ommer.
\newblock {High-Resolution Image Synthesis with Latent Diffusion Models}.
\newblock In \emph{Proceedings of the IEEE/CVF Conference on Computer Vision and Pattern Recognition (CVPR)}, pages 10684--10695, 2022{\natexlab{b}}.

\bibitem[Ruiz et~al.(2023)Ruiz, Li, Jampani, Pritch, Rubinstein, and Aberman]{Dreambooth}
Nataniel Ruiz, Yuanzhen Li, Varun Jampani, Yael Pritch, Michael Rubinstein, and Kfir Aberman.
\newblock {DreamBooth: Fine Tuning Text-to-Image Diffusion Models for Subject-Driven Generation}.
\newblock In \emph{Proceedings of the IEEE/CVF conference on computer vision and pattern recognition}, pages 22500--22510, 2023.

\bibitem[Safaee et~al.(2024)Safaee, Mikaeili, Patashnik, Cohen-Or, and Mahdavi-Amiri]{Clic}
Mehdi Safaee, Aryan Mikaeili, Or Patashnik, Daniel Cohen-Or, and Ali Mahdavi-Amiri.
\newblock {CLiC: Concept Learning in Context}.
\newblock In \emph{Proceedings of the IEEE/CVF Conference on Computer Vision and Pattern Recognition}, pages 6924--6933, 2024.

\bibitem[Sangkloy et~al.(2017)Sangkloy, Lu, Fang, Yu, and Hays]{Scribbler}
Patsorn Sangkloy, Jingwan Lu, Chen Fang, Fisher Yu, and James Hays.
\newblock {Scribbler: Controlling Deep Image Synthesis with Sketch and Color}.
\newblock In \emph{Proceedings of the IEEE conference on computer vision and pattern recognition}, pages 5400--5409, 2017.

\bibitem[Schonberger and Frahm(2016)]{Structure}
Johannes~L Schonberger and Jan-Michael Frahm.
\newblock {Structure-From-Motion Revisited}.
\newblock In \emph{Proceedings of the IEEE conference on computer vision and pattern recognition}, pages 4104--4113, 2016.

\bibitem[Siyao et~al.(2021)Siyao, Zhao, Yu, Sun, Metaxas, Loy, and Liu]{Atd12k}
Li Siyao, Shiyu Zhao, Weijiang Yu, Wenxiu Sun, Dimitris Metaxas, Chen~Change Loy, and Ziwei Liu.
\newblock {Deep Animation Video Interpolation in the Wild}.
\newblock In \emph{Proceedings of the IEEE/CVF conference on computer vision and pattern recognition}, pages 6587--6595, 2021.

\bibitem[Song et~al.(2020)Song, Meng, and Ermon]{Ddim}
Jiaming Song, Chenlin Meng, and Stefano Ermon.
\newblock {Denoising Diffusion Implicit Models}.
\newblock \emph{arXiv preprint arXiv:2010.02502}, 2020.

\bibitem[Song et~al.(2023)Song, Zhang, Lin, Cohen, Price, Zhang, Kim, and Aliaga]{Objectstitch}
Yizhi Song, Zhifei Zhang, Zhe Lin, Scott Cohen, Brian Price, Jianming Zhang, Soo~Ye Kim, and Daniel Aliaga.
\newblock {ObjectStitch: Object Compositing with Diffusion Model}.
\newblock In \emph{Proceedings of the IEEE/CVF Conference on Computer Vision and Pattern Recognition}, pages 18310--18319, 2023.

\bibitem[Su et~al.(2020)Su, Chu, and Huang]{instance-aware}
Jheng-Wei Su, Hung-Kuo Chu, and Jia-Bin Huang.
\newblock {Instance-Aware Image Colorization}.
\newblock In \emph{Proceedings of the IEEE/CVF conference on computer vision and pattern recognition}, pages 7968--7977, 2020.

\bibitem[Sun and Wu(2019)]{Layout_style}
Wei Sun and Tianfu Wu.
\newblock {Image Synthesis From Reconfigurable Layout and Style}.
\newblock In \emph{Proceedings of the IEEE/CVF International Conference on Computer Vision}, pages 10531--10540, 2019.

\bibitem[Sun and Wu(2021)]{Layout_style2}
Wei Sun and Tianfu Wu.
\newblock {Learning Layout and Style Reconfigurable Gans for Controllable Image Synthesis}.
\newblock \emph{IEEE Transactions on Pattern Analysis and Machine Intelligence}, 44\penalty0 (9):\penalty0 5070--5087, 2021.

\bibitem[Sushko et~al.(2020)Sushko, Sch{\"o}nfeld, Zhang, Gall, Schiele, and Khoreva]{Gan2}
Vadim Sushko, Edgar Sch{\"o}nfeld, Dan Zhang, Juergen Gall, Bernt Schiele, and Anna Khoreva.
\newblock {You Only Need Adversarial Supervision for Semantic Image Synthesis}.
\newblock \emph{arXiv preprint arXiv:2012.04781}, 2020.

\bibitem[Tang et~al.(2023)Tang, Jia, Wang, Phoo, and Hariharan]{Dift}
Luming Tang, Menglin Jia, Qianqian Wang, Cheng~Perng Phoo, and Bharath Hariharan.
\newblock {Emergent Correspondence from Image Diffusion}.
\newblock \emph{Advances in Neural Information Processing Systems}, 36:\penalty0 1363--1389, 2023.

\bibitem[Tang et~al.(2024)Tang, Ruiz, Chu, Li, Holynski, Jacobs, Hariharan, Pritch, Wadhwa, Aberman, et~al.]{Realfill}
Luming Tang, Nataniel Ruiz, Qinghao Chu, Yuanzhen Li, Aleksander Holynski, David~E Jacobs, Bharath Hariharan, Yael Pritch, Neal Wadhwa, Kfir Aberman, et~al.
\newblock {RealFill: Reference-Driven Generation for Authentic Image Completion}.
\newblock \emph{ACM Transactions on Graphics (TOG)}, 43\penalty0 (4):\penalty0 1--12, 2024.

\bibitem[Tumanyan et~al.(2023)Tumanyan, Geyer, Bagon, and Dekel]{Plug-and-play}
Narek Tumanyan, Michal Geyer, Shai Bagon, and Tali Dekel.
\newblock {Plug-and-Play Diffusion Features for Text-Driven Image-to-Image Translation}.
\newblock In \emph{Proceedings of the IEEE/CVF Conference on Computer Vision and Pattern Recognition}, pages 1921--1930, 2023.

\bibitem[Utintu et~al.(2024)Utintu, Chowdhury, Sain, Koley, Bhunia, and Song]{Sketchdeco}
Chaitat Utintu, Pinaki~Nath Chowdhury, Aneeshan Sain, Subhadeep Koley, Ayan~Kumar Bhunia, and Yi-Zhe Song.
\newblock {SketchDeco: Decorating B\&W Sketches with Colour}.
\newblock \emph{arXiv preprint arXiv:2405.18716}, 2024.

\bibitem[Wang et~al.(2024{\natexlab{a}})Wang, Chen, and Wang]{wang2024diffusion}
Bingyuan Wang, Qifeng Chen, and Zeyu Wang.
\newblock {Diffusion-Based Visual Art Creation: A Survey and New Perspectives}.
\newblock \emph{arXiv preprint arXiv:2408.12128}, 2024{\natexlab{a}}.

\bibitem[Wang et~al.(2025)Wang, Meng, Cao, Cai, Li, Ma, Chen, and Wang]{wang2025magicscroll}
Bingyuan Wang, Hengyu Meng, Rui Cao, Zeyu Cai, Lanjiong Li, Yue Ma, Qifeng Chen, and Zeyu Wang.
\newblock {MagicScroll: Enhancing Immersive Storytelling with Controllable Scroll Image Generation}.
\newblock In \emph{2025 IEEE Conference Virtual Reality and 3D User Interfaces (VR)}. IEEE, 2025.

\bibitem[Wang et~al.(2024{\natexlab{b}})Wang, Ma, Guo, Xiao, Huang, and Li]{wang2024cove}
Jiangshan Wang, Yue Ma, Jiayi Guo, Yicheng Xiao, Gao Huang, and Xiu Li.
\newblock {COVE: Unleashing the Diffusion Feature Correspondence for Consistent Video Editing}.
\newblock \emph{arXiv preprint arXiv:2406.08850}, 2024{\natexlab{b}}.

\bibitem[Wang et~al.(2024{\natexlab{c}})Wang, Pu, Qi, Guo, Ma, Huang, Chen, Li, and Shan]{wang2024taming}
Jiangshan Wang, Junfu Pu, Zhongang Qi, Jiayi Guo, Yue Ma, Nisha Huang, Yuxin Chen, Xiu Li, and Ying Shan.
\newblock {Taming Rectified Flow for Inversion and Editing}.
\newblock \emph{arXiv preprint arXiv:2411.04746}, 2024{\natexlab{c}}.

\bibitem[Wang et~al.(2018)Wang, Liu, Zhu, Tao, Kautz, and Catanzaro]{Cgan}
Ting-Chun Wang, Ming-Yu Liu, Jun-Yan Zhu, Andrew Tao, Jan Kautz, and Bryan Catanzaro.
\newblock {High-Resolution Image Synthesis and Semantic Manipulation with Conditional Gans}.
\newblock In \emph{Proceedings of the IEEE conference on computer vision and pattern recognition}, pages 8798--8807, 2018.

\bibitem[Wang et~al.(2019)Wang, Jabri, and Efros]{Cycleconsistency}
Xiaolong Wang, Allan Jabri, and Alexei~A Efros.
\newblock {Learning Correspondence From the Cycle-Consistency of Time}.
\newblock In \emph{Proceedings of the IEEE/CVF conference on computer vision and pattern recognition}, pages 2566--2576, 2019.

\bibitem[Wang et~al.(2004)Wang, Bovik, Sheikh, and Simoncelli]{Imagequalityassessment}
Zhou Wang, Alan~C Bovik, Hamid~R Sheikh, and Eero~P Simoncelli.
\newblock {Image Quality Assessment: From Error Visibility to Structural Similarity}.
\newblock \emph{IEEE Transactions on Image Processing}, 13\penalty0 (4):\penalty0 600--612, 2004.

\bibitem[Wu et~al.(2023)Wu, Yang, Xu, Liu, Yan, and Zhang]{Flexicon}
Shukai Wu, Yuhang Yang, Shuchang Xu, Weiming Liu, Xiao Yan, and Sanyuan Zhang.
\newblock {FlexIcon: Flexible Icon Colorization via Guided Images and Palettes}.
\newblock In \emph{Proceedings of the 31st ACM International Conference on Multimedia}, pages 8662--8673, 2023.

\bibitem[Xie et~al.(2023)Xie, Zhao, Xiao, Chan, Li, Xu, Zhang, and Hou]{Dreaminpainter}
Shaoan Xie, Yang Zhao, Zhisheng Xiao, Kelvin~CK Chan, Yandong Li, Yanwu Xu, Kun Zhang, and Tingbo Hou.
\newblock {DreamInpainter: Text-Guided Subject-Driven Image Inpainting With Diffusion Models}.
\newblock \emph{arXiv preprint arXiv:2312.03771}, 2023.

\bibitem[Xu et~al.(2024)Xu, Zhang, Liew, Yan, Liu, Zhang, Feng, and Shou]{Magicanimate}
Zhongcong Xu, Jianfeng Zhang, Jun~Hao Liew, Hanshu Yan, Jia-Wei Liu, Chenxu Zhang, Jiashi Feng, and Mike~Zheng Shou.
\newblock {MagicAnimate: Temporally Consistent Human Image Animation Using Diffusion Model}.
\newblock In \emph{Proceedings of the IEEE/CVF Conference on Computer Vision and Pattern Recognition}, pages 1481--1490, 2024.

\bibitem[Xue et~al.(2024)Xue, Wang, Tian, Ma, Wang, Zhao, Min, Zhao, Zhang, Shum, et~al.]{xue2024follow}
Jingyun Xue, Hongfa Wang, Qi Tian, Yue Ma, Andong Wang, Zhiyuan Zhao, Shaobo Min, Wenzhe Zhao, Kaihao Zhang, Heung-Yeung Shum, et~al.
\newblock {Follow-Your-Pose v2: Multiple-Condition Guided Character Image Animation for Stable Pose Control}.
\newblock \emph{arXiv preprint arXiv:2406.03035}, 2024.

\bibitem[Yan et~al.(2023)Yan, Ito, Moriai, and Saito]{Twosteptraining}
Dingkun Yan, Ryogo Ito, Ryo Moriai, and Suguru Saito.
\newblock {Two-Step Training: Adjustable Sketch Colourization via Reference Image and Text Tag}.
\newblock In \emph{Computer Graphics Forum}, page e14791. Wiley Online Library, 2023.

\bibitem[Yan et~al.(2024)Yan, Yuan, Wu, Nishioka, Fujishiro, and Saito]{Colorizediffusion}
Dingkun Yan, Liang Yuan, Erwin Wu, Yuma Nishioka, Issei Fujishiro, and Suguru Saito.
\newblock {Colorizediffusion: Adjustable Sketch Colorization With Reference Image and Text}.
\newblock \emph{arXiv preprint arXiv:2401.01456}, 2024.

\bibitem[Yang et~al.(2023)Yang, Gu, Zhang, Zhang, Chen, Sun, Chen, and Wen]{Paintbyexample}
Binxin Yang, Shuyang Gu, Bo Zhang, Ting Zhang, Xuejin Chen, Xiaoyan Sun, Dong Chen, and Fang Wen.
\newblock {Paint by Example: Exemplar-Based Image Editing with Diffusion Models}.
\newblock In \emph{Proceedings of the IEEE/CVF Conference on Computer Vision and Pattern Recognition}, pages 18381--18391, 2023.

\bibitem[Yuan et~al.(2024{\natexlab{a}})Yuan, Huang, He, Ge, Shi, Chen, Luo, and Yuan]{yuan2024identity}
Shenghai Yuan, Jinfa Huang, Xianyi He, Yunyuan Ge, Yujun Shi, Liuhan Chen, Jiebo Luo, and Li Yuan.
\newblock {Identity-Preserving Text-to-Video Generation by Frequency Decomposition}.
\newblock \emph{arXiv preprint arXiv:2411.17440}, 2024{\natexlab{a}}.

\bibitem[Yuan et~al.(2024{\natexlab{b}})Yuan, Huang, Shi, Xu, Zhu, Lin, Cheng, Yuan, and Luo]{yuan2024magictime}
Shenghai Yuan, Jinfa Huang, Yujun Shi, Yongqi Xu, Ruijie Zhu, Bin Lin, Xinhua Cheng, Li Yuan, and Jiebo Luo.
\newblock {MagicTime: Time-Lapse Video Generation Models as Metamorphic Simulators}.
\newblock \emph{arXiv preprint arXiv:2404.05014}, 2024{\natexlab{b}}.

\bibitem[Zabari et~al.(2023)Zabari, Azulay, Gorkor, Halperin, and Fried]{DiffusingColors}
Nir Zabari, Aharon Azulay, Alexey Gorkor, Tavi Halperin, and Ohad Fried.
\newblock {Diffusing Colors: Image Colorization with Text Guided Diffusion}.
\newblock In \emph{SIGGRAPH Asia 2023 Conference Papers}, pages 1--11, 2023.

\bibitem[Zhang et~al.(2019)Zhang, He, Liao, Sander, Yuan, Bermak, and Chen]{example1}
Bo Zhang, Mingming He, Jing Liao, Pedro~V Sander, Lu Yuan, Amine Bermak, and Dong Chen.
\newblock {Deep Exemplar-Based Video Colorization}.
\newblock In \emph{Proceedings of the IEEE/CVF conference on computer vision and pattern recognition}, pages 8052--8061, 2019.

\bibitem[Zhang et~al.(2018)Zhang, Li, Wong, Ji, and Liu]{Twostagesketch}
Lvmin Zhang, Chengze Li, Tien-Tsin Wong, Yi Ji, and Chunping Liu.
\newblock {Two-stage Sketch Colorization}.
\newblock \emph{ACM Transactions on Graphics (TOG)}, 37\penalty0 (6):\penalty0 1--14, 2018.

\bibitem[Zhang et~al.(2023)Zhang, Rao, and Agrawala]{controlnet}
Lvmin Zhang, Anyi Rao, and Maneesh Agrawala.
\newblock {Adding Conditional Control to Text-to-Image Diffusion Models}.
\newblock In \emph{Proceedings of the IEEE/CVF International Conference on Computer Vision}, pages 3836--3847, 2023.

\bibitem[Zhang et~al.(2021)Zhang, Wang, Wen, Li, and Liu]{automaticanimation}
Qian Zhang, Bo Wang, Wei Wen, Hai Li, and Junhui Liu.
\newblock {Line Art Correlation Matching Feature Transfer Network for Automatic Animation Colorization}.
\newblock In \emph{Proceedings of the IEEE/CVF Winter Conference on Applications of Computer Vision}, pages 3872--3881, 2021.

\bibitem[Zhang et~al.(2016{\natexlab{a}})Zhang, Isola, and Efros]{Colorfulimagecolor}
Richard Zhang, Phillip Isola, and Alexei~A Efros.
\newblock {Colorful Image Colorization}.
\newblock In \emph{Computer Vision--ECCV 2016: 14th European Conference, Amsterdam, The Netherlands, October 11-14, 2016, Proceedings, Part III 14}, pages 649--666. Springer, 2016{\natexlab{a}}.

\bibitem[Zhang et~al.(2016{\natexlab{b}})Zhang, Isola, and Efros]{scrible3}
Richard Zhang, Phillip Isola, and Alexei~A Efros.
\newblock {Colorful Image Colorization}.
\newblock In \emph{Computer Vision--ECCV 2016: 14th European Conference, Amsterdam, The Netherlands, October 11-14, 2016, Proceedings, Part III 14}, pages 649--666. Springer, 2016{\natexlab{b}}.

\bibitem[Zhang et~al.(2024)Zhang, Huang, Ma, Li, Luo, Xie, Qin, Luo, Li, Liu, et~al.]{Ram}
Youcai Zhang, Xinyu Huang, Jinyu Ma, Zhaoyang Li, Zhaochuan Luo, Yanchun Xie, Yuzhuo Qin, Tong Luo, Yaqian Li, Shilong Liu, et~al.
\newblock {Recognize Anything: A Strong Image Tagging Model}.
\newblock In \emph{Proceedings of the IEEE/CVF Conference on Computer Vision and Pattern Recognition}, pages 1724--1732, 2024.

\bibitem[Zhao et~al.(2021)Zhao, Wu, Liu, and He]{example2}
Hengyuan Zhao, Wenhao Wu, Yihao Liu, and Dongliang He.
\newblock {Color2Embed: Fast Exemplar-Based Image Colorization Using Color Embeddings}.
\newblock \emph{arXiv preprint arXiv:2106.08017}, 2021.

\bibitem[Zhao et~al.(2020)Zhao, Han, Shao, and Snoek]{pixelated}
Jiaojiao Zhao, Jungong Han, Ling Shao, and Cees~GM Snoek.
\newblock {Pixelated Semantic Colorization}.
\newblock \emph{International Journal of Computer Vision}, 128:\penalty0 818--834, 2020.

\bibitem[Zhu et~al.(2024{\natexlab{a}})Zhu, Li, Ma, He, and Li]{zhu2024multibooth}
Chenyang Zhu, Kai Li, Yue Ma, Chunming He, and Xiu Li.
\newblock {MultiBooth: Towards Generating All Your Concepts in an Image from Text}.
\newblock \emph{arXiv preprint arXiv:2404.14239}, 2024{\natexlab{a}}.

\bibitem[Zhu et~al.(2024{\natexlab{b}})Zhu, Li, Ma, Tang, Fang, Chen, Chen, and Li]{zhu2024instantswap}
Chenyang Zhu, Kai Li, Yue Ma, Longxiang Tang, Chengyu Fang, Chubin Chen, Qifeng Chen, and Xiu Li.
\newblock {InstantSwap: Fast Customized Concept Swapping across Sharp Shape Differences}.
\newblock \emph{arXiv preprint arXiv:2412.01197}, 2024{\natexlab{b}}.

\end{thebibliography}
}

\end{document}